\documentclass{article}

\PassOptionsToPackage{numbers, sort&compress}{natbib}

\usepackage[dblblindworkshop,final]{neurips_2026}
\workshoptitle{Child Safety in AI}

\usepackage[utf8]{inputenc}
\usepackage[T1]{fontenc}
\usepackage{hyperref}
\hypersetup{hidelinks,pdftitle={A Rank-One Bypass of Fixed Linear Erasure},pdfauthor={Jai Kumar Sharma, Vanshaj Khattar}}
\usepackage{url}
\usepackage{booktabs}
\usepackage{array}
\usepackage{amsmath,amssymb,amsfonts}
\usepackage{amsthm}
\theoremstyle{definition}
\newtheorem{readertheorem}{Theorem}
\usepackage{graphicx}
\usepackage{microtype}
\usepackage{placeins}

\title{A Rank-One Bypass of Fixed Linear Erasure}

\author{Jai Kumar Sharma \qquad Vanshaj Khattar \\
Virginia Polytechnic Institute and State University}

\begin{document}

\maketitle

\begin{abstract}
Linear concept erasure can make a target unreadable to a specified family of linear classifiers applied to a model's current internal features. In an open-weight release, however, an attacker can modify the layers that produce those features. We study a fixed affine erasure layer whose upstream model is allowed to change. If the layer transmits even one direction visible to an allowed downstream classifier, the attacker can write a target score into that direction and carry it through the layer. We show that this condition is also necessary, and that for a nonconstant linear target score the required change is a rank-one affine update.
Across six CIFAR-10 class-pair tasks, the route restores nearly all pre-erasure target performance and can leave a deployed benign linear head's scores unchanged to numerical tolerance. We then price the attack: with eight target labels and \(2{,}000\) unlabelled calibration examples, a \(10\%\) mean relative feature edit yields \(79.8\%\) test accuracy, which is \(91.2\%\) of the fitted score's gain over chance; tighter edit limits sharply reduce success. Linear guardedness therefore remains a valid guarantee about the current representation, but it is not by itself a tamper-resistance guarantee for an editable open-weight model.
\end{abstract}

\section{Introduction}
\label{sec:intro}

Open-weight image generators pose a distinct child-safety release problem. Once model weights are distributed, developers can no longer enforce hosted prompt filters, output classifiers, or rate limits. Locally run and fine-tuned models have been associated with AI-generated child sexual abuse material (CSAM) \citep{iwf2023ai,thiel2023genml,iwf2026harm}, and CSAM has been identified in training data behind widely redistributed checkpoints \citep{thiel2023laion}. These risks motivate safety interventions applied before release \citep{thorn2024sbd}.

Concept erasure is one such intervention. Existing methods edit diffusion-model weights or remove target directions from internal representations \citep{gandikota2023erasing,kumari2023ablating,gandikota2024unified}. We study fixed affine erasure of representations, with LEACE as the canonical example, because its guarantee can be stated exactly. On the distribution used to fit it and for the stated convex loss, LEACE constructs a minimal-distortion transform under which no predictor in a chosen linear family outperforms a constant predictor \citep{belrose2023leace}. This is a precise guarantee about what linear classifiers can read from the model's current features. It is not, and LEACE does not claim it to be, a guarantee about what the released model can compute after an attacker changes the layers that produce those features.

We therefore ask a release-level question: if the erasure layer remains fixed, can an attacker modify the upstream model and route the target around it? Prior work already shows that nonlinear probes can decode linearly erased representations and that concept edits can reverse under later adaptation \citep{ravfogel2022linear,ravfogel2022kernelized,pham2024circumventing,suriyakumar2026unstable,liu2025erased}. \citet{cretu2026evaluating} further measure cost to success in a child-safety filtering game. We do not claim nonlinear decoding or fine-tuning reversal as new; we isolate a different mechanism: the attacker computes a target score upstream and writes it into a direction that the fixed affine layer still transmits.

\paragraph{Contributions.}
\emph{(i) Condition.} For a fixed affine layer \(E(z)=Az+c\) and allowed linear-head space \(H\),
a route exists exactly when \(P_HA\neq0\); for a nonconstant linear target score it is a
rank-one affine adapter. The condition is elementary; we measure what it costs.
\emph{(ii) Constrained construction.} In all six evaluated pairs, the route can leave a deployed
benign linear head's scores unchanged, with drift at most \(1.2\times10^{-10}\).
\emph{(iii) Cost.} With eight target labels, a \(10\%\) mean relative feature edit recovers
\(0.912\) of the target score's gain over chance, whereas a \(1\%\) edit recovers \(0.25\).
\emph{(iv) Audits.} A weak probe can overstate erasure, and fine-tuning around the frozen layer
recovers the target as fast as with no erasure, within a \(0.05\) margin.

\paragraph{Scope.}
We distinguish \emph{readout} of the unchanged \(E(x)\), \emph{routing} through a changed input to
\(E\), and \emph{adaptation} of a larger upstream block; Theorem~\ref{thm:rank1} concerns routing
through a fixed affine layer, and nonlinear, noisy, or refitted bottlenecks are outside it. Our
experiments use benign class-pair proxies in discriminative image models, so we make no direct
quantitative claim about diffusion models, CSAM-domain data, or harmful generation.

\begin{figure}[t]
\centering
\includegraphics[width=0.97\linewidth]{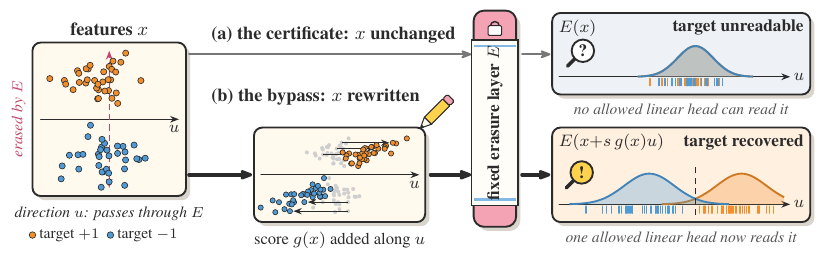}
\caption{\textbf{A representation certificate is not a release certificate.}
Schematic with simulated two-dimensional points: \(E\) erases the dashed vertical direction, which carries the target, and transmits the horizontal direction \(u\).
\textbf{(a)} With the features unchanged, no allowed linear head can read the target from \(E(x)\).
\textbf{(b)} An open-weight attacker edits the upstream model to add a target score \(g(x)\) along \(u\); the same fixed layer transmits it, and an allowed linear head recovers the target. Such a route exists exactly when \(P_HA\neq0\); for nonconstant linear \(g\), it is a rank-one affine adapter.}
\label{fig:overview}
\end{figure}

\section{Threat model: fixed eraser, editable upstream}
\label{sec:scope}

We analyze the released pipeline
\(
\text{input}\to x\to E(x)=Ax+c\to h^\top E(x)+\beta
\).
All weights are public. In the restricted attack studied here, the adversary may modify the layers producing \(x\) and choose any downstream linear classifier (\emph{head}) \(h\in H\) and bias \(\beta\), but must leave \(E\) bit-identical. An unrestricted attacker could delete \(E\), so success under this restriction is a conservative failure case.

Each target attack fits a score \(g(x)\) from \(N\) labelled examples and uses \(2{,}000\) unlabelled target-domain examples only to calibrate the edit size. The attack protocol is fixed in advance on held-out pairs; the one-time pilot and complete resource accounting are reported separately in Appendix~\ref{app:frontier} and Table~\ref{tab:resource}.

\section{One surviving direction is enough}
\label{sec:theorem}

The attacker does not need to recover the target from the unchanged erased representation. Instead, it computes a numerical target score before the erasure layer and writes that score into one direction \(u\) that the fixed layer still transmits (Figure~\ref{fig:overview}b):
\begin{equation}
  r_s(x)
  =
  x+s\,g(x)u ,
  \label{eq:route}
\end{equation}
where \(g(x)\) is the target score and \(s\) controls the route's strength. The score may depend on every coordinate of \(x\), but the added displacement always lies in the one-dimensional space spanned by \(u\).

Choose the downstream bias \(\beta=-h^\top c\) to cancel the fixed offset of \(E(z)=Az+c\). An allowed head \(h\) then sees
\begin{equation}
\begin{aligned}
  h^\top E(r_s(x))+\beta
  &=
  \underbrace{h^\top Ax}_{\text{original downstream term}}
  +
  \underbrace{s\,(h^\top Au)\,g(x)}_{\text{routed target score}} .
\end{aligned}
\label{eq:routed-logit}
\end{equation}

If \(h^\top Au\neq0\), the attacker can flip the sign of \(u\) so that \(h^\top Au>0\). Increasing \(s\) then makes the routed score dominate the original downstream term. The layer therefore does not need to retain the target in the original \(E(x)\); it only needs to transmit one direction after the upstream representation has been changed. LEACE constrains a readout of the current \(E(x)\), not a later rewrite of \(x\).

\begin{readertheorem}[One surviving direction is enough]
\label{thm:rank1}
Let \(E(z)=Az+c\) be fixed, and let the allowed heads form a subspace \(H\) with orthogonal projector \(P_H\). Suppose the editable upstream model can compute a target score \(g(x)\) that takes both signs. There exist a direction \(u\), a head \(h\in H\), and a bias \(\beta\) such that, for every \(x\) with \(g(x)\neq0\), the decision of \(h\) on \(E(r_s(x))\) equals \(\operatorname{sign}g(x)\) for all sufficiently large \(s\), if and only if \(P_HA\neq0\). For unrestricted heads, this reduces to \(A\neq0\). For nonconstant linear \(g\), the update \(r_s\) is a rank-one affine adapter. If \(P_HA=0\), every allowed head has an input-independent output.
\end{readertheorem}

\noindent\emph{Proof sketch.}
In Equation~\eqref{eq:routed-logit}, the routed copy of \(g\) has gain \(\kappa=h^\top Au\), which can be made positive exactly when \(P_HA\neq0\); it fixes the sign of the output once \(s>|h^\top Ax|/(\kappa|g(x)|)\). The finite-sample statement and proof are in Appendix~\ref{app:theorem}. Rescaling \(A\) by a positive constant does not help: it scales the original and routed terms equally.

\section{Does the route work, and what does it cost?}
\label{sec:cost}

\paragraph{Does the analytic route work?}
We evaluate six CIFAR-10 class-pair tasks using ImageNet-pretrained ResNet-18 features behind a LEACE layer, with five seeds per pair. Once the route strength exceeds the threshold above on every evaluation point, the analytic rank-one adapter recovers the accuracy of the pre-erasure target score, which ranges from \(0.87\) to \(1.00\). The head-only baseline and the constant-map control \(A=0\) remain at each task's base rate, as predicted by Theorem~\ref{thm:rank1}. Trained rank-one adapters match the closed-form construction in every pair--seed run (Appendix~\ref{app:theorem}; Figure~\ref{fig:theorem}).

\paragraph{Can it preserve deployed benign scores?}
Suppose a deployed benign linear head with class-score vectors \(h_1,\ldots,h_q\) also reads the representation. In our experiments it is a softmax head over the eight non-target CIFAR-10 classes (\(q=8\); Appendix~\ref{app:frontier}). In all six pairs, a route that preserves its scores exists, because eight constraints remove at most eight of the \(511\) directions that \(A\) transmits. We choose \(u\) so that
\begin{equation}
  h_j^\top Au=0,
  \qquad
  j=1,\ldots,q .
  \label{eq:protected-logits}
\end{equation}
In exact arithmetic, Equation~\eqref{eq:protected-logits} leaves every benign score unchanged for every input and every route strength. In our implementation, the maximum observed drift is \(1.2\times10^{-10}\), and the benign head's predictions are identical to those of the clean model. The constrained route remains within \(0.3\) percentage points of the unconstrained route on every run. By contrast, the unconstrained route reduces this head's accuracy from \(0.81\)--\(0.87\) to \(0.27\)--\(0.71\). We call the constrained route \emph{score-preserving}.

\begin{figure}[t]
\centering
\includegraphics[width=0.97\linewidth]{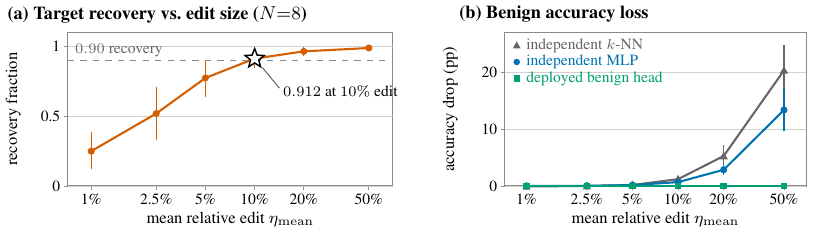}
\caption{\textbf{What the route costs, and what its exact guarantee covers.}
\textbf{(a)}~With eight labels, recovery rises with the mean relative edit, from \(0.25\) at \(1\%\) to \(0.912\) at the reported \(10\%\) point (star); the dashed line marks \(0.90\).
\textbf{(b)}~The deployed benign head's accuracy is unchanged at every edit size, whereas independent MLP and \(k\)-NN evaluators lose accuracy as the edit grows. Error bars are \(95\%\) hierarchical-bootstrap intervals over six pairs and five seeds.}
\label{fig:frontier}
\end{figure}

\paragraph{How expensive is it?}
We fit the target score \(g_N\) from \(N\) labelled examples and limit the average feature change using
\begin{equation}
  \eta_{\mathrm{mean}}
  =
  \frac{\mathbb E\|r_s(x)-x\|_2}
       {\mathbb E\|x\|_2}.
  \label{eq:edit-budget}
\end{equation}
This is a mean relative edit, not a per-example maximum.

At eight labels and a \(10\%\) mean relative edit (95th-percentile edit \(21\%\)), the route reaches \(0.798\) test accuracy and recovers \(0.912\,[0.843,0.976]\) of the target score's gain over chance, pooled over runs (Figure~\ref{fig:frontier}a). This is an aggregate rather than a per-run guarantee: 20 of 30 pair--seed runs exceed \(0.90\) recovery, while uniform \(0.90\) recovery requires a \(50\%\) mean edit. At eight labels, pooled recovery falls to \(0.52\) at a \(2.5\%\) mean edit and \(0.25\) at a \(1\%\) mean edit. Thus, bounding the allowed feature change materially raises the cost of this route.

\paragraph{Controls and broader benign effects.}
Replacing the target-score direction with a random one while keeping the routing direction \(u\) fixed reduces attack accuracy by \(0.24\) absolute \([0.20,0.30]\), and fitting the same procedure to permuted labels returns performance to approximately chance. Target-aligned score information, rather than routing geometry alone, therefore drives recovery. At the \(10\%\) operating point, independent benign MLP and \(k\)-NN accuracy fall by \(0.7\) and \(1.2\) percentage points, respectively, with larger losses under larger edits (Figure~\ref{fig:frontier}b). Exact preservation of selected linear scores is therefore not preservation of the full representation.

\section{Why erasure evaluations need strong probes and matched controls}
\label{sec:audits}

\paragraph{Probe strength and numerical precision matter.}
\label{sec:precision}
A weak probe can make erasure look stronger than it is. A linear head trained by gradient descent reads approximately chance (\(\approx0.51\)) from LEACE-erased ResNet-18 features, whereas a closed-form ridge solver reaches \(0.83\)--\(0.96\) on the same features when fit and solve both use single precision. The cause is rounding error, not routing: at every other precision we tested, including double, the same solver returns to approximately \(0.51\). Erasers should therefore be fitted in double precision and audited with a closed-form probe. Separately, an eraser fitted in double precision still leaves \(0.585\)--\(0.598\) accuracy on held-out data: suppression on the fitting sample need not transfer to new data (Appendix~\ref{app:ridge}).

\paragraph{Fine-tuning recovery needs matched controls.}
\label{sec:relearning}
A frozen erasure layer adds no measurable delay to fine-tuning (Appendix~\ref{app:relearning}). Fine-tuning one upstream block around it reaches \(0.917\) target accuracy after ten steps, but the undefended model and a control with a random rank-one edit in place of the erasure adapt at the same rate: all differences lie within a \(0.05\) equivalence margin fixed in advance, at every sampled checkpoint (the first is step \(10\)). A reinitialized block is slower, so pretrained features matter. Unlike the score-preserving route, this attack is costly to benign use: deployed benign-head accuracy falls from approximately \(0.84\) to \(0.39\). It also differs from the analytic route in labels and parameters (Table~\ref{tab:resource}), so we do not rank them.

\section{What does linear erasure certify?}
\label{sec:discussion}

Linear guardedness is a valid guarantee about a fixed representation and a stated classifier family;
it is not by itself evidence of robust capability removal from an editable release. Our cost results
cut both ways: tight edit limits sharply weaken this route, but an open-weight release has no
mechanism to enforce them, so an edit limit is an evaluation budget, not a defense. Open-weight
evaluations should therefore report an adaptive-attacker cost-to-success frontier. Within the affine
model, only a layer that transmits nothing to the allowed heads blocks the route
(Theorem~\ref{thm:rank1}); whether nonlinear, noisy, or clipped bottlenecks resist it is open.

\paragraph{Limitations.}
We use benign class-pair proxies and fixed affine layers: six CIFAR-10 pairs that share classes, one
fine-tuning budget, two small nonlinear evaluators, and no control pretrained without the target
classes, so generic and target-specific pretraining are not separated. \emph{Guarded under the
evaluated attacker class} is therefore more precise than \emph{removed}, although erasure may still
serve as defense in depth. A benign generative proxy \citep{cretu2026evaluating} is the natural
next test.

\clearpage

\appendix

\section*{Ethics statement}
We use only benign public image-classification datasets and do not access, generate, store, or
display CSAM-domain material. The child-safety relevance is methodological: we test what a fixed
linear-erasure certificate supports under open-weight modification, not harmful-generation
performance.

\section{Formal routing result}
\label{app:theorem}

\paragraph{Setting.}
The released model computes \(h^\top E(x)+\beta\), where \(x\in\mathbb{R}^d\) is an editable
representation and
\[
  E(z)=Az+c,
  \qquad
  A\in\mathbb{R}^{m\times d},\quad c\in\mathbb{R}^m,
\]
is fixed. Allowed downstream linear classifiers (heads) lie in a subspace
\(H\subseteq\mathbb{R}^m\), with orthogonal projector \(P_H\), and the bias \(\beta\) is free.
Let \(t\in\{-1,+1\}\) be the target label and let
\(g:\mathbb{R}^d\to\mathbb{R}\) be any target score the attacker can evaluate upstream. Define
\[
  a^\star=\Pr[\operatorname{sign}g(x)=t].
\]

\paragraph{Full form of Theorem~\ref{thm:rank1}.}
An allowed one-dimensional route exists if and only if \(P_HA\neq0\).
When this condition holds, choose \(u\in\mathbb{R}^d\) and \(h\in H\) so that
\(\kappa:=h^\top Au>0\), and set \(\beta=-h^\top c\). For
\[
  r_s(x)=x+s\,g(x)u,
  \qquad
  \rho(x)=h^\top Ax,
\]
the attacked head output is
\begin{equation}
  h^\top E(r_s(x))+\beta
  =
  \rho(x)+s\kappa g(x).
  \label{eq:app-routed-logit}
\end{equation}
Consequently:

\begin{enumerate}
  \item If \(g(x)\neq0\) and
  \[
    s>\frac{|\rho(x)|}{\kappa |g(x)|},
  \]
  then the attacked decision equals \(\operatorname{sign}g(x)\).

  \item On a finite evaluation set \(\{x_i\}_{i=1}^n\) with \(g(x_i)\neq0\), define
  \[
    s^\dagger
    =
    \max_{i:\,\operatorname{sign}\rho(x_i)\neq\operatorname{sign}g(x_i)}
    \frac{|\rho(x_i)|}{\kappa |g(x_i)|},
  \]
  with \(s^\dagger=0\) if the disagreement set is empty. Every \(s>s^\dagger\) reproduces
  \(\operatorname{sign}g\) on all points; \(s^\dagger\) is the sharp infimum threshold, with
  equality depending on the tie rule.

  \item If \(\Pr[g(x)=0]=0\), then
  \[
    \Pr[\operatorname{sign}(h^\top E(r_s(x))+\beta)=t]
    \longrightarrow a^\star
    \qquad\text{as }s\to\infty.
  \]

  \item If \(g(x)=w^\top x+b\) is nonconstant and linear, then
  \[
    r_s(x)
    =
    (I+suw^\top)x+sbu,
  \]
  whose linear perturbation has rank one.

  \item If \(P_HA=0\), then \(h^\top Au=0\) for every \(u\) and every \(h\in H\), so no route of
  this form can transmit an input-dependent score to an allowed head.
\end{enumerate}

\begin{proof}
Equation~\eqref{eq:app-routed-logit} follows from affine linearity and
\(\beta=-h^\top c\). If \(s\kappa|g(x)|>|\rho(x)|\), the routed term has larger magnitude than
the original term and fixes the sign, proving the pointwise and finite-set statements. For
asymptotic agreement, let
\[
  D_s(x)
  =
  \mathbf{1}\!\left\{
  \operatorname{sign}(\rho(x)+s\kappa g(x))
  \neq
  \operatorname{sign}g(x)
  \right\}.
\]
For every \(x\) with \(g(x)\neq0\), \(D_s(x)\to0\), and \(0\le D_s\le1\). Dominated convergence
therefore gives \(\Pr[D_s=1]\to0\), which implies convergence of target accuracy to \(a^\star\).
The linear-\(g\) form is obtained by substitution. Finally,
\[
  \max_{\|u\|=\|h\|=1,\,h\in H} h^\top Au
  =
  \sigma_{\max}(P_HA),
\]
so a positive gain exists exactly when \(P_HA\neq0\).
\end{proof}

\paragraph{Scale-to-edit conversion.}
The route changes the representation by \(s\,g(x)u\), hence
\begin{equation}
  \eta_{\mathrm{mean}}(s)
  =
  \frac{\mathbb{E}\|r_s(x)-x\|_2}{\mathbb{E}\|x\|_2}
  =
  \frac{s\|u\|_2\,\mathbb{E}|g(x)|}{\mathbb{E}\|x\|_2}.
  \label{eq:app-scale-to-edit}
\end{equation}
The experiments fix \(\eta_{\mathrm{mean}}\) and solve
Equation~\eqref{eq:app-scale-to-edit} for \(s\) on an unlabelled calibration pool.

\paragraph{Score-preserving route.}
Let \(h_1,\ldots,h_q\) be the class-score vectors of the deployed benign head and define
\[
  U_{\mathrm{safe}}
  =
  \bigcap_{j=1}^q \ker(h_j^\top A),
\]
with orthogonal projector \(\Pi_{\mathrm{safe}}\). A route that leaves every benign
score unchanged exists exactly when
\begin{equation}
  P_HA\Pi_{\mathrm{safe}}\neq0.
  \label{eq:app-safe-condition}
\end{equation}
Indeed, \(u\in U_{\mathrm{safe}}\) gives \(h_j^\top Au=0\) for every \(j\), while the
maximum allowed target gain is
\[
  \sigma_{\max}(P_HA\Pi_{\mathrm{safe}}).
\]

\paragraph{Boundary cases.}
For unrestricted heads, \(P_H=I\), so only the constant map \(A=0\) blocks the route. Replacing
\(A\) by \(\lambda A\), \(\lambda>0\), scales the original and routed terms equally and therefore
does not change the required \(s\) in the noiseless affine sign-decision model. Clipping, additive
noise, quantization, saturation, and dynamic refitting are outside this result.

\paragraph{Six-pair verification.}
Across all six CIFAR-10 pairs and five seeds, the analytic route recovers the direct
target-score accuracy, which ranges from \(0.868\) to \(0.996\). The head-only baseline
reads \(0.50\)--\(0.51\), and the constant-map control \(A=0\) remains at the task base
rate. A separately trained rank-one adapter matches the closed-form construction in every
evaluated pair--seed run.

\begin{figure}[t]
\centering
\includegraphics[width=\linewidth]{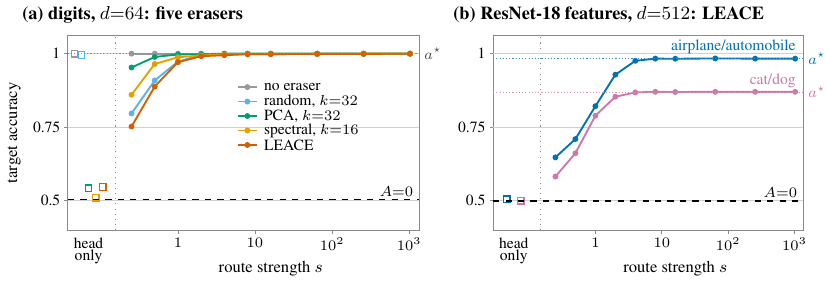}
\caption{\textbf{One surviving direction is enough (Theorem~1).} Target accuracy of the allowed head versus route strength $s$ for the analytic rank-one route: \textbf{(a)}~digits behind five erasers, \textbf{(b)}~ResNet-18 features behind LEACE. Every nonzero eraser reaches $a^\star$ (dotted); only $A{=}0$ (dashed) stays at chance. Open squares: the head without a route. Panel~(b) shows two of the six pairs, with $a^\star=0.982$ (airplane/automobile) and $0.868$ (cat/dog).}
\label{fig:theorem}
\end{figure}

\paragraph{Digits check (Figure~\ref{fig:theorem}a).}
As a small second setting we use the scikit-learn \(8\times8\) digits data, standardized
(\(d=64\)), with four digit pairs (0/1, 3/8, 5/6, 2/7) and two seeds each. The five fixed maps are
the identity (no eraser), removal of a random \(32\)-dimensional subspace, removal of the top
\(32\) principal components, removal of a \(16\)-dimensional subspace spanned by differences of
class-conditional cluster means (``spectral''), and LEACE; \(A=0\) is the control. Each curve
averages the eight pair--seed runs.

\FloatBarrier
\section{Cost-frontier protocol, controls, and resources}
\label{app:frontier}

\paragraph{Protocol.}
We evaluate six CIFAR-10 class pairs and five seeds per pair. Each of the resulting \(30\) runs
sweeps
\[
  N\in\{8,16,32,64,128,256,512,1000,4000\}
\]
labelled examples for fitting \(g_N\), and
\[
  \eta\in\{0.01,0.025,0.05,0.10,0.20,0.50\}.
\]
Splits are class-balanced and disjoint: \(4{,}000\) examples fit the eraser, \(4{,}000\) form the
attacker pool, \(2{,}000\) unlabelled examples calibrate the route strength, and the untouched
CIFAR-10 test split is read once. Label budgets are nested balanced prefixes. The attack family,
regularizer, sign convention, and threshold are fixed in advance on two held-out class pairings;
the resulting one-time \(10{,}720\)-label protocol-design cost is reported separately from each
target attack.

The route is calibrated by solving Equation~\eqref{eq:app-scale-to-edit} on the unlabelled pool.
At the highlighted point, calibration sets \(\eta_{\mathrm{mean}}=0.10\); the realized test mean is
\(0.1002\), and test \(\eta_{95}=0.210\). Table~\ref{tab:frontier-summary} lists selected points of the frontier and
Table~\ref{tab:frontier-pairs} the per-pair results at the highlighted point.

\paragraph{Metrics.}
For balanced binary tasks, let \(a_N^\star\) be the direct accuracy of \(g_N\), and let
\(a_{\mathrm{route}}\) be the routed accuracy. We report
\[
  \mathrm{RF}
  =
  \frac{a_{\mathrm{route}}-0.5}{a_N^\star-0.5},
\]
the fraction of \(g_N\)'s advantage over chance transmitted by the route. As a robustness check we
also use
\[
  \mathrm{RF}_{\mathrm{inc}}
  =
  \frac{a_{\mathrm{route}}-a_{\mathrm{head}}}
       {a_N^\star-a_{\mathrm{head}}},
\]
which normalizes against the head-only accuracy \(a_{\mathrm{head}}\) of the released model.

Uncertainty intervals resample class pairs and then seeds within each sampled pair. They
summarize variability over the selected six-pair, five-seed panel rather than an unrestricted
population of concepts.

\begin{table}[!htb]
\caption{\textbf{Selected points from the measured route frontier.}
At \(N=8,\eta=0.10\), routed test accuracy is \(0.798\). The highlighted point summarizes
the reported test frontier and was not confirmed on separate data. All \(30\) runs reach
\(\mathrm{RF}\ge0.90\) only at \(\eta=0.50\).}
\label{tab:frontier-summary}
\centering
\small
\setlength{\tabcolsep}{5pt}
\begin{tabular}{@{}cccc@{}}
\toprule
Labels \(N\) & Mean edit \(\eta\) & Pooled \(\mathrm{RF}\) & Runs with \(\mathrm{RF}\ge0.90\) \\
\midrule
8 & 0.01  & \(0.250\,[0.119,0.385]\) & \(0/30\) \\
8 & 0.025 & \(0.519\,[0.327,0.713]\) & \(3/30\) \\
8 & 0.10  & \(\mathbf{0.912}\,[0.843,0.976]\) & \(20/30\) \\
\bottomrule
\end{tabular}
\end{table}

\begin{table}[!htb]
\caption{\textbf{Pair-level heterogeneity at the highlighted operating point.}
Values above one mean that, within seed variation, the routed classifier matches or slightly exceeds
the direct test reading of \(g_N\).}
\label{tab:frontier-pairs}
\centering
\small
\setlength{\tabcolsep}{5pt}
\begin{tabular}{@{}lcc@{}}
\toprule
Class pair & Mean \(\mathrm{RF}\) at \(N=8,\eta=0.10\) & Runs \(\ge0.90\) \\
\midrule
airplane / automobile & 0.835 & \(2/5\) \\
automobile / truck    & 0.781 & \(1/5\) \\
bird / deer           & 1.001 & \(5/5\) \\
cat / dog             & 0.961 & \(4/5\) \\
dog / horse           & 0.936 & \(4/5\) \\
frog / ship           & 0.963 & \(4/5\) \\
\bottomrule
\end{tabular}
\end{table}

\paragraph{Deployed benign head.}
For each target pair, the deployed benign task is \(8\)-way classification of the eight CIFAR-10
classes outside the pair (base rate \(0.125\)). Its head is a single linear softmax layer on clean
\(E(x)\), trained on \(8{,}000\) images of those classes with the cross-entropy loss using Adam at
learning rate \(0.05\) for \(1{,}500\) full-batch steps, and then frozen. Its eight class-score
vectors are \(h_1,\ldots,h_8\), so the constraint removes at most eight of the
\(511\) directions that \(A\) transmits. Clean test accuracy on the eight classes is
\(0.81\)--\(0.87\) across runs.

\paragraph{Controls and benign effects.}
The score-preserving route differs from the unconstrained route by at most
\(0.003\) absolute target accuracy in every pair--seed run. By contrast, the
unconstrained route reduces the deployed benign head's accuracy to \(0.27\)--\(0.71\).
The independent evaluators are a one-hidden-layer width-\(128\) ReLU MLP trained on
clean \(E(x)\), and Euclidean \(k\)-NN with \(k=10\). The MLP is trained with Adam at
learning rate \(0.003\) for \(1{,}200\) full-batch steps under the cross-entropy loss and then
frozen. Neither evaluator is used to construct or select the routing direction.
Replacing only the target-score direction while keeping the routing direction \(u\) fixed reduces
accuracy by \(0.24\,[0.20,0.30]\) absolute. Fitting the same procedure to permuted labels returns
performance to approximately chance. The benign scores move by at most
\(4.3\times10^{-13}\) over the full edit grid, with identical benign-head predictions; at the much larger route strengths used to verify Theorem~\ref{thm:rank1}, the worst-case drift is \(1.2\times10^{-10}\). At
\(\eta=0.10\), independent benign MLP and \(k\)-NN evaluators lose
\(0.68\,[0.44,0.94]\) and \(1.21\,[0.73,1.75]\) percentage points, respectively.

At \(N=8,\eta=0.10\),
\[
  \mathrm{RF}_{\mathrm{inc}}
  =
  0.905\,[0.833,0.971],
\]
close to the chance-normalized value because the head-only accuracy is \(0.510\).

\paragraph{Construction-specific frontier.}
The reported surface prices the closed-form route obtained from the top singular pair of
\(P_HA\Pi_{\mathrm{safe}}\). It is not a proof of the globally cheapest rank-one attack.

\begin{table}[!htb]
\caption{\textbf{Attack resource vector at the highlighted operating point.}
The supported claim is that, after fitting the small score \(g_N\), the routing layer is constructed
with zero routing-optimization steps. The fine-tuning baseline was not timed in the same
environment, so no wall-clock speedup is claimed. The fine-tuning benign-head loss is measured at step \(10\), the first sampled
post-start checkpoint. Appendix~\ref{app:relearning} reports the later
deployed benign-head decline from approximately \(0.84\) to \(0.39\).}
\label{tab:resource}
\centering
\small
\setlength{\tabcolsep}{4.5pt}
\begin{tabular}{@{}p{6.0cm}cc@{}}
\toprule
Resource & Analytic route & Last-block fine-tuning \\
\midrule
Target training labels & \(8\) & \(4{,}000\) \\
Target validation labels & \(0\) & \(2{,}000\) \\
Unlabelled calibration examples & \(2{,}000\) & -- \\
Held-out-pair pilot labels (one-time) & \(10{,}720\) & -- \\
Installed parameters & \(1{,}538\) & \(8{,}394{,}241\) \\
Gradient-optimized parameters & \(513\) & \(8{,}394{,}241\) \\
Analytically set parameters & \(1{,}025\) & \(0\) \\
Optimizer steps & \(2{,}000\) on \(N\) rows & \(\ge10\) on \(4{,}000\) images \\
Routing optimization after fitting \(g_N\) & \(0\) & n/a \\
Linear-algebra solve & one \(512\times512\) SVD & -- \\
\(\eta_{\mathrm{mean}}/\eta_{95}\) & \(0.100/0.210\) & unconstrained \\
Benign-score drift at \(N=8,\eta=0.10\) & \(\le2.4\times10^{-13}\) & not constrained \\
Independent-evaluator accuracy loss & \(0.7\) pt MLP, \(1.2\) pt \(k\)-NN & not measured \\
Deployed benign-head accuracy loss at step \(10\) & \(0\) pt & \(11\)--\(17\) pt \\
\bottomrule
\end{tabular}
\end{table}

\FloatBarrier
\section{Solver, numerical precision, and transfer to new data}
\label{app:ridge}

\begin{table}[!htb]
\caption{\textbf{Summary of the linear-solver audit on LEACE-erased ResNet-18 features.}
The first five rows vary fit and solve arithmetic jointly. The last row is a separate held-out
generalization experiment.}
\label{tab:precision-summary}
\centering
\small
\setlength{\tabcolsep}{5pt}
\begin{tabular}{@{}lcc@{}}
\toprule
Evaluation setting & Gradient-trained head & Closed-form linear attacker \\
\midrule
float32 fit and solve & \(\approx0.51\) & \(0.83\)--\(0.96\) \\
float64 fit and solve & \(\approx0.51\) & \(\approx0.508\) \\
TF32 fit and solve & \(\approx0.51\) & \(\approx0.508\) \\
bf16 fit and solve & \(\approx0.51\) & \(\approx0.508\) \\
fp16 fit and solve & \(\approx0.51\) & \(\approx0.508\) \\
float64 fit, disjoint held-out data & --              & \(0.585\)--\(0.598\) \\
\bottomrule
\end{tabular}
\end{table}

\paragraph{Effect of arithmetic precision.}
A native-float32 LEACE fit leaves a coherent near-null direction that a scale-insensitive
closed-form solve can amplify. Float64 removes the fitted empirical direction more accurately,
whereas TF32, bf16, and fp16 disrupt the small coherent residue through coarser rounding. These
are different mechanisms. Because fitting and solving precision were varied together, the supported
claim is an end-to-end arithmetic effect, not a storage-only effect.

\paragraph{Held-out transfer.}
The exact-float64 held-out result is separate: empirical guardedness on the eraser-fit sample does
not transfer perfectly to new data. This does not contradict LEACE's stated fitted-distribution
guarantee. Table~\ref{tab:precision-summary} reports the audited settings supporting the
main-text comparison.

\subsection{Theorem-relevant loss versus thresholded accuracy}
\label{app:metric}

LEACE's formal guarantee concerns convex risk, not thresholded accuracy. On the eraser-fit split,
the best audited linear predictor has no excess predictive value over the constant baseline under
the theorem-relevant loss. On held-out splits the excess predictive value is positive, matching the
held-out accuracy result above. The distinction is therefore between a fitted-distribution risk
certificate and behavior on new samples, not a contradiction of the guardedness theorem.

\FloatBarrier
\section{Fine-tuning controls and equivalence analysis}
\label{app:relearning}

\paragraph{Design.}
Four arms share \texttt{conv1..layer3} frozen and \texttt{layer4} plus a fresh head trainable, for
the same \(8{,}394{,}241\)-parameter budget: (A) undefended identity, (B) frozen LEACE,
(C) a concept-blind random rank-one edit, and (D) a reinitialized last block. All arms use Adam,
learning rate \(10^{-4}\), batch size \(128\), and \(4{,}000\) labelled images. A disjoint
\(2{,}000\)-example labelled validation split is used only to monitor target accuracy at each
checkpoint; it is not used for gradient updates, for selecting the learning rate or a reported
checkpoint (the learning rate is fixed and every sampled checkpoint is reported), or for the final
test metric. We evaluate six
class pairs, three seeds, and steps
\(\{0,10,25,50,100,200,400\}\) (Table~\ref{tab:rca}). Uncertainty is obtained by resampling pairs and then seeds.

\begin{table}[!htb]
\caption{\textbf{Fine-tuning controls.}
Target accuracy is averaged over six pairs and three seeds. LEACE differs from the undefended and
random-edit arms by at most \(0.005\) at every sampled checkpoint.}
\label{tab:rca}
\centering
\scriptsize
\setlength{\tabcolsep}{3.2pt}
\begin{tabular}{@{}lrrrrrrr@{}}
\toprule
 & 0 & 10 & 25 & 50 & 100 & 200 & 400 \\
\midrule
A undefended target acc.   & .497 & .915 & .947 & .957 & .964 & .966 & .966 \\
B LEACE target acc.        & .502 & .917 & .947 & .957 & .964 & .966 & .966 \\
C random rank-1 target acc.& .496 & .916 & .947 & .957 & .964 & .966 & .965 \\
D reinitialized target acc.& .507 & .794 & .852 & .930 & .943 & .951 & .957 \\
\midrule
B--A                       & +.005 & +.002 & 0.000 & 0.000 & 0.000 & 0.000 & 0.000 \\
B--C                       & +.005 & +.001 & 0.000 & 0.000 & 0.000 & 0.000 & 0.000 \\
B--D                       & -.006 & +.123 & +.095 & +.027 & +.021 & +.015 & +.008 \\
\bottomrule
\end{tabular}
\end{table}

\paragraph{Equivalence result.}
Using a \(0.05\) accuracy margin fixed in advance, the LEACE arm is equivalent to the undefended
and random-edit controls at the sampled checkpoints; one-sided bootstrap upper bounds on the
per-run maximum gap are \(0.005\) and \(0.004\), respectively. The first post-start checkpoint is
step \(10\), so a transient confined to steps \(1\)--\(9\) is not excluded.

The reinitialized arm is slower: the LEACE-minus-reinitialized difference is
\(0.123\,[0.066,0.189]\) at step \(10\) and
\(0.008\,[0.002,0.018]\) at step \(400\). This shows that retained pretrained representation
quality matters, but the present controls do not separate generic from target-specific pretraining;
a concept-naive pretraining control would be required for that distinction.

The large fine-tuning attack is not score-preserving: deployed benign-head accuracy falls from
approximately \(0.84\) to \(0.39\).

\end{document}